# Hydroelectric Generation Forecasting with Long Short Term Memory (LSTM) Based Deep Learning Model for Turkey

Mehmet BULUT,[1,2,3, a]

**AFFILIATIONS**
[1]Electricity Generation Inc., General Directorate, Ankara, TURKEY
[2]Atilim University, School of Civil Aviation, Ankara, TURKEY
[3]mehmetbulut06@gmail.com
[a]Orcid. 0000-0003-3998-1785

**ABSTRACT**

Hydroelectricity is one of the renewable energy source, has been used for many years in Turkey. The production of hydraulic power plants based on water reservoirs varies based on different parameters. For this reason, the estimation of hydraulic production gains importance in terms of the planning of electricity generation. In this article, the estimation of Turkey's monthly hydroelectricity production has been made with the long-short-term memory (LSTM) network-based deep learning model. The designed deep learning model is based on hydraulic production time series and future production planning for many years. By using real production data and different LSTM deep learning models, their performance on the monthly forecast of hydraulic electricity generation of the next year has been examined. The obtained results showed that the use of time series based on real production data for many years and deep learning model together is successful in long-term prediction. In the study, it is seen that the 100-layer LSTM model, in which 120 months (10 years) hydroelectric generation time data are used according to the RMSE and MAPE values, are the highest model in terms of estimation accuracy, with a MAPE value of 0.1311 (13.1%) in the annual total and 1.09% as the monthly average distribution. In this model, the best results were obtained for the 100-layer LSTM model, in which the time data of 144 months (12 years) hydroelectric generation data are used, with a RMSE value of 29,689 annually and 2474.08 in monthly distribution. According to the results of the study, time data covering at least 120 months of production is recommended to create an acceptable hydropower forecasting model with LSTM.

*Keywords:* Long Short Term Memory (LSTM), Deep Learning, Hydroelectric power, Generation Forecasting.

**1. INTRODUCTION**

Renewable energy can meet two-thirds of total global energy demand and contribute to the massive reduction of greenhouse gas emissions needed between now and 2050 to limit average global surface temperature rise below 2°C [1]. Examples of major renewable energy sources are solar, wind, hydrogen, hydroelectric, wave and geothermal. The biggest environmental advantages of renewable energy sources are that they are renewable due to their continued existence in nature, they do not harm nature by reducing carbon emissions, and they are clean and sustainable energy sources [2]. Renewable energy resources such as sunlight, wind speed and monthly average total precipitation change seasonally. Renewable energy are mainly dependent on local environmental conditions such as temperature and precipitation-runoff rates [3]. Accurate estimation of hydroelectric power generation is a critical issue for the current power management process, and it is of great benefit in planning future electricity generation, optimizing the generation and thus ensuring system integration [4]. Due to the volatile nature of renewable energy, power generation from this resources is highly volatile, making the calculation and estimation of the supply to the electricity grid is critical. In this sense, the development of deep learning technology has achieved remarkable results in many fields, especially in computer vision and prediction. In applications, deep learning models are created based on time series data, many of which are applied to real life such as precipitation-flow modeling [5]. Deep learning architectures allow the learning construct to learn complex relationships between inputs and outputs and complex patterns in data [6]. Long Short Term Memory (LSTM) networks are attracting renewed attention today and are replacing many practical applications of time series forecasting systems [7]. Wang et al. examined the applicable data preprocessing techniques and post-error correction methods to improve the efficiency, efficiency and application potential of deep learning-based renewable energy forecasting methods and forecasting accuracy [8].

Cheng et al., to achieve efficiency and reliability of power demand forecasting in smart grid in terms of demand response and resource allocation [9], forecasting energy demand 3 days in advance during each month of the year [10], non-linear, non-stationary series of short-term electric load time series. and due to its non-seasonal nature [11], to design a new custom power demand forecasting algorithm based on deep learning method on end-user power demand models [12], last 24 hours consumption data and such as temperature, humidity, wind speed and radiation. Long Short-Term Memory (LSTM) in studies aimed at estimating the consumption value for the next hour using weather data [13], producing hourly forecasts up to 72 hours with long-term wind speed and power forecasting based on meteorological information for a wind park [14]. They propose a power demand forecasting model based on a neural network. Li et al. tried improving the prediction performance based on deep learning by fusing different production time data components such as daily, weekly and long time of a hydraulic power plant [15]. Del Real et al., on the other hand, used a mixed deep learning architecture consisting of a convolutional neural network (CNN) combined with an artificial neural network (ANN) to perform energy demand estimation [16].

Hydroelectric generation is one of Turkey's leading renewable energy sources [17]. However, electricity generation from hydraulic power plants is completely dependent on water resources and production differs from month to month. A deep learning approach is proposed that is applied directly to the collected data to predict anomalies, such as when the power demand load is higher than the available power [18]. Load estimation is performed using convolutional Neural Network components to extract distinct useful features from the historical load data array [19]. To create a prediction model with Turkey's monthly electricity generation dataset, methods such as LSTM based on a deep learning algorithm were used using time series [20]. Renewable energy sources such as solar, wind and hydroelectricity mainly depend on local environmental and meteorological conditions such as temperature and precipitation-runoff rates. In this article, it is developed a model based on a deep learning algorithm by using time series to create a forecast model with Turkey's monthly hydroelectric electricity generation dataset instead of local environmental and meteorological conditions such as temperature and precipitation-flow rates in the monthly medium-term estimation of the amount of energy produced by Hydroelectric Power Plants (HPPs).

In this study is used a dataset covering monthly hydroelectric production values of Turkey in the period January 2007 – December 2018, using monthly hydroelectric generation information, is distributed monthly with LSTM networks. A one-year hydroelectric generation forecasting system has been developed. LSTM Network-Based Deep Learning Model and Monthly Hydroelectric Generation Forecasting System were developed for developing both production values and annual hydroelectric installed capacity values. The dataset used in the study was divided into three parts and estimation was conducted with three different LSTM models. From the data obtained at the end of the study, it was seen that the Short-Long-Term Memory network-based forecasting system was successful within the acceptance limits by using the time series containing the hydroelectric generation data and the installed power values.

## 2. HYDROLECTRICITY GENERATION IN TURKEY

Electricity generation from renewable sources such as wind, solar and geothermal, especially hydro-electricity, shows a rapid increase every year in Turkey. By the end of 2019, Turkey's installed electricity capacity increased by 3.07% compared to 2018 and reached 91,267 MW [21]. As of the end of 2019, 48.7% of the total installed power consisted of power plants producing from renewable energy sources, and their total installed power capacity reached 44,405 MW. With 20,642,5 MW of installed power, dam hydraulic power plants and 7,860.5 MW were formed by river power plants. As of the end of 2019, Turkey's installed power and gross electricity generation reached 304,252 GWh. In 2019, 29.2% of Turkey's electricity generation comes from hydroelectric sources, 7% from wind energy, 3.55% from solar energy, 2.7% from geothermal energy and 1.5% from other sources. provided. The average annual precipitation in Turkey is around 642.6 mm, which corresponds to 501 billion m3 of water per year.

However, within the framework of today's technical and economic conditions, the surface water potential that can be consumed for various purposes is an average of 98 billion m3 per year, of which 95 billion m3 is from local streams and 3 billion m3 is from rivers from neighboring countries. Turkey's theoretical hydroelectric potential was calculated as 433 billion kWh, technical potential as 216 billion kWh and both technically and economically viable potential as 130 billion kWh. Turkey's gross potential is 1% of the world's total potential (433,000 GWh/year) and 16% of Europe's total. It shows the increase in the installed power of hydraulic power plants in Turkey. As can be seen, there is a regular increase in the hydraulic energy installed power. As shown in Figure 1, Turkey's hydroelectric production varies in some years, showing a continuous upward trend in installation and generation, consistent with total electricity generation [22].

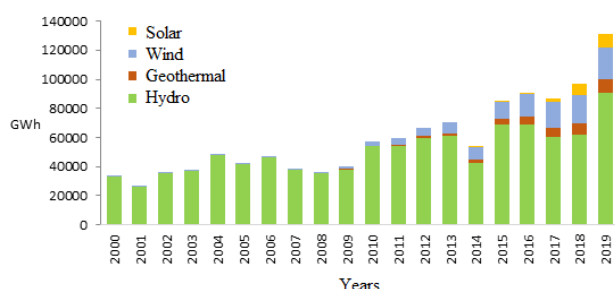

**Figure 1.** Electricity generation from various renewable sources including dam type hydro.

Considering the 12-year production covering the hydroelectric production data of Turkey between 2007–

2018, the minimum, maximum and average production values of the productions corresponding to the months between January and December are given in Figure 2. It is seen that the highest production was realized in May, and the lowest production was in October.

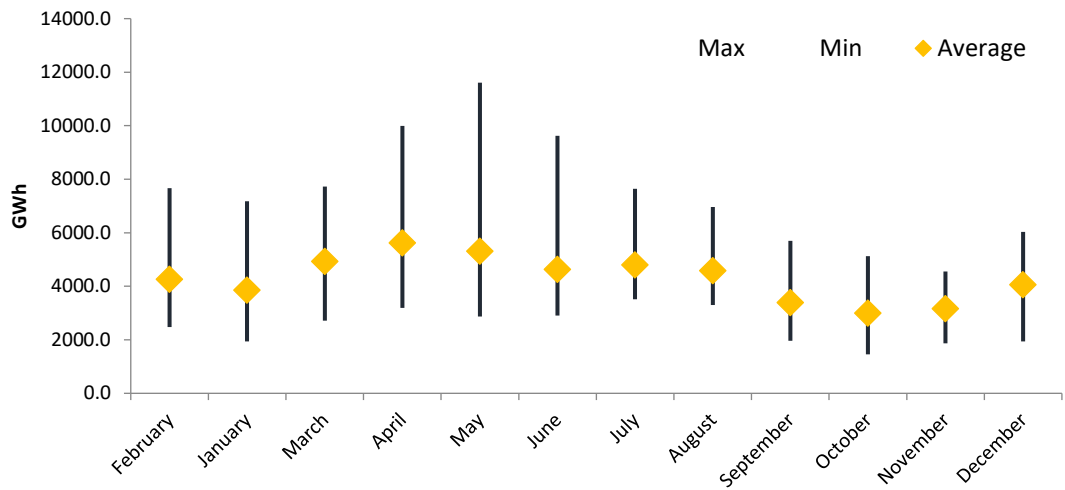

**Figure 2.** Monthly Maximum-Minimum-Average values of hydroelectric generation for many years

## 3. DEEP LEARNING WITH LSTM NETWORKS

Artificial Neural networks artificially imitate the nature and functioning of human neural networks. ANN is widely used for special applications such as pattern recognition, data classification and so on. A trained feedforward neural network can be used to predict the output from an inference. Deep learning (DL) has solved many complex problems of artificial intelligence (AI) that have existed for many years. In fact, DL models are deeper variants of multiple-layer artificial neural networks (ANNs), whether linear or non-linear. Each layer is connected to its lower and upper layers with different weights. The ability of DL models to learn hierarchical properties from various data types such as numerical, image, text, and audio makes them powerful in recognizing, regressing, solving semi-supervised and unsupervised problems [23]. For example, it can predict whether the image is a square or an equilateral rectangle, as shown in Figure 3 below. The operation of the two networks operates independently of each other and does not carry a memory to hold the information during the previous operation.

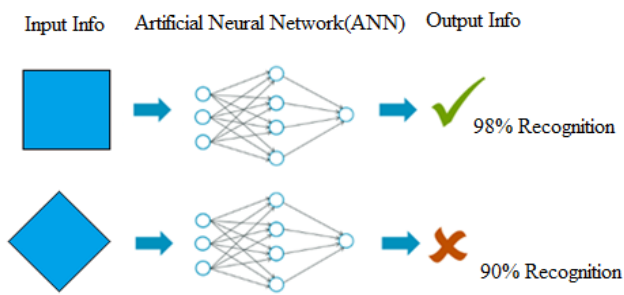

**Figure 3.** Pattern recognition ANN's capability an by understanding the difference between shapes

But there are many situations where prior data knowledge is important due to some problems. Such as: handwriting recognition, voice recognition, etc. They do not contain memory units in traditional ANN structures. Therefore, these networks do not have the memory to understand sequential data such as language translation. Deep neural network (DNN) is one of the most common DL models that includes multiple layers of linear and nonlinear operations.

### A. Structure of long short-term memory networks

LSTMs can remember information for a long time. Many variations have been developed to solve the problem posed in a deep Recursive Neural Network (RNN). To solve this problem, LSTM uses gate units to decide what information to keep or remove from the previous state [25]. Conceptually, the repeating unit of an LSTM cell tries to "remember" all the past information the network has ever seen and "forget" irrelevant data. These different purposes are done by adding different activation function layers called "gates" to the network structure. A LSTM Network consists of three different Gates, they are Forget gate, Input gate and Output gate. Gates control the flow of information from memory to memory. The gates are controlled by combining the output from the previous time step and the current input, and optionally the cell state vector [26].

**Cell State Vector:** Represents the memory of the LSTM and undergoes changes through forgetting of old memory (forget gate) and the addition of new memory (gateway).

**Input Gate ($i$):** It determines the scope of the information to be written to the Internal Cell Status. The Sigmoid Function controls what new information is added to the cell state from the Current input. Decides how much of this unit to add to the current state.

**Forget gate ($f$):** Determines to what extent previous data will be forgotten. The Sigmoid Function controls what information is discarded from Memory. It decides how much of the past you should remember.

**Output Gate ( $o$ ):** Determines which output (next Hidden State) will be produced from the current Internal Cell State. The Sigmoid Function conditionally decides what to remove from Memory. Decides which part of the current cell to output.

**Modulation function ( $g$ ):** Hyperbolic Tangent Function. It is considered to be a sub-part of the Gateway. It is used to modulate the information that the input gate will write to the Internal State Cell by adding nonlinearity to the information and Zero-averaging the information.

The basic workflow of a Long Short-Term Memory Network is similar to that of a Recurrent Neural Network; the only difference is that the Inner Cell State is also transmitted along with the Hidden State (Figure 4).

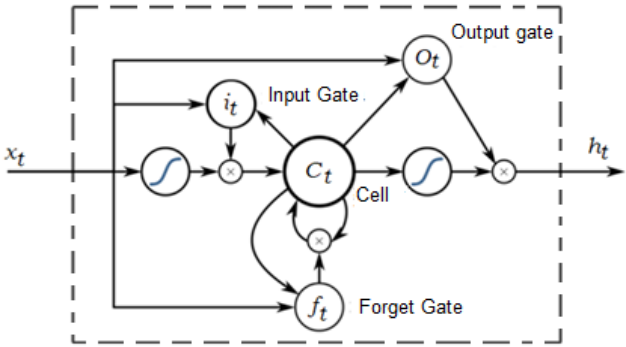

**Figure 4.** Gate connections of an LSTM cell

The core component of LSTMs is a memory cell that can hold information over time controlled by gates. It can maintain its state over time, consisting of an open memory (also called a cell state vector) and gate units. Gate units regulate the flow of information entering and leaving the memory.

The input gate is used to check whether the state in the current cell allows it to be overridden by external information. The entrance gate in the LST cell structure used in the study is shown in Equation (1).

$$i_t = \sigma_k(W_i x_t + U_i h_{t-1} + b_i) \quad (1)$$

Here;
$i_t$ : entrance gate vector,
$\sigma_k$ : sigmoid function,
$x_t$: input vector,
$W_i$ and $U_i$ : parameter matrices
$b_i$: bias vector.

The output vector decides whether to keep the state in the current cell, which will affect other cells, and is defined as in the equation below.

$$o_t = \sigma_k(W_o x_t + U_o h_{t-1} + b_o) \quad (2)$$

Another cell gate defined in the LSTM memory cell is the forgotten container, which allows the LSTM to reset its state and is defined as follows.

$$f_t = \sigma_k(W_f x_t + U_f h_{t-1} + b_f) \quad (3)$$

As a result, equations 4 and 5 below show how the cell state and the output vector are obtained from the input gate, forget gate and output gate.

$$c_t = f_t \odot c_{t-1} + i_t \odot tanh(W_f x_t + U_c h_{t-1} + b_c) \quad (4)$$

$$h_t = o_t \odot \sigma_h(c_t) \text{ or } h_t = o_t \odot tanh(c_t) \quad (5)$$

Here, $\odot$ represents the Hadamart product, $\sigma_c$ and $\sigma_h$ represents the hyperbolic tangent functions.

## 4. HYDRO GENERATION FORECASTING

The hydroelectric generation estimation study consists of the following steps. In this section, the hydroelectric generation estimation framework made with LSTM is given in Figure 6.

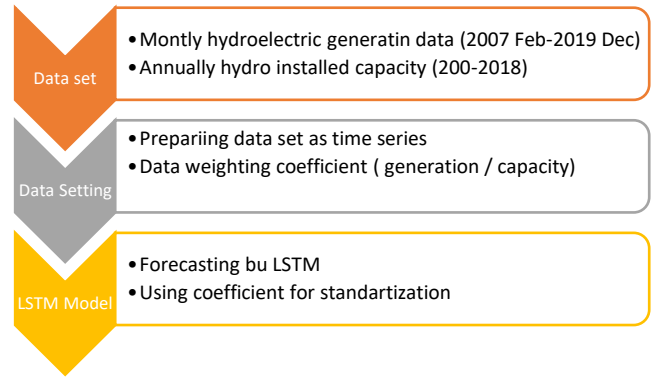

**Figure 6.** LSTM hydropower generation forecasting framework

We can express the definition of the problem in the hydroelectric production estimation study as follows:

1- Obtaining a hydroelectric generation time series by using the monthly total electricity generation values realized in hydroelectric power plants for many years.
2- Using this obtained time series, forecasting annual hydroelectric generation monthly in the future.
3- Extract the internal feature to help better predict hydropower generation: Using the 12-year hydropower generation, the monthly correction coefficient is obtained by obtaining the average hydropower generation for each month of the year during January-December and standardizing them in a max-min range.
4- Historical observations of previous t - 1 time steps $X_{t-1} = \{X_1, X_2, \cdots, X_{t-1}\}$ and internal features D(y): January..December, considering the hydroelectric generation forecast, is to learn a model that predicts electricity generation value. Time t, that is, in time step t of the electricity generation value $X_t$ {Xt | Estimating Xt <1, Ft}.

$$X_{(n)}(t) = Xn(1), Xn(2), ...., Xn(t-1), ..., Xn+1(1), Xn+1(2),....,Xn+1(t-1),...X_N(1), X_N(2),....,X_N(t-1) \quad (5)$$

Here;
   n=1..N, N=12 (January,…,December)
   t = number of past years,
   X(n+1)(t) = n in year t. hydroelectric production of the month.

## A. Specifying Input Data for LSTM Network

The production of a dam hydroelectric power station is affected by keeping the water level too high or too low due to the change in water flow. Similarly, the generation of river or canal hydroelectric power plants depends on the water regime affected by precipitation. The most important parameter showing the generation characteristics of hydroelectric power plants is the capacity factor. The capacity factor shows how many hours the hydraulic plant produces in 8765 hours of the year and is used for performance comparison. Net plant capacity factor (CF) is referred to as full capacity plant can produce energy part of the total energy generated in a given period.

$$CF = ANP / ( IP *365 \text{ days} *24 \text{ hours/day} ) \quad (6)$$

Here,
ANP: Annual Electricity Production
IP: Installed Capacity

Hydroelectric power plants, on the other hand, have an average capacity factor of around 40%. I.e., about 3500 corresponding to the maximum duration time of the year 40% of the year of manufacture can be produced. In this situation; Since the hydroelectric production data used in this study covers a period of 13 years, the installed capacity value that provides the production for each year is not the same due to each hydraulic installed capacity increase. To save the hydroelectric generation time data from this installed capacity value change, the time series based on monthly generation performance was obtained using the following formula (Figure 8 a, b).

$$GP = MEP/IP \quad (7)$$

Here,
GP: Generation Performance
MEP: Monthly Electricity Production

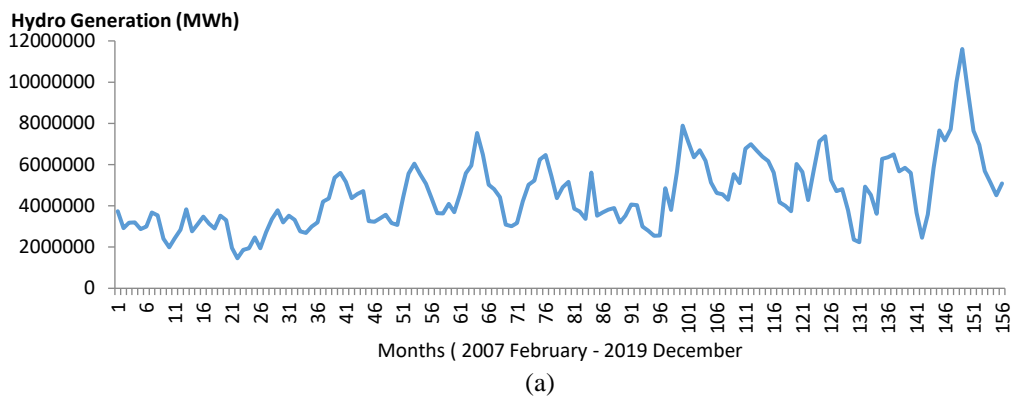

(a)

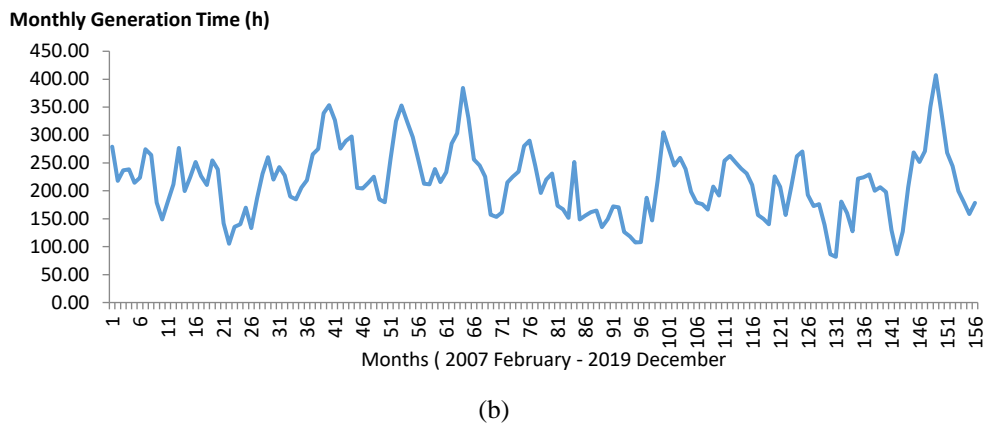

(b)

**Figure 8**. a) Monthly Hydroelectric production 2007–2019 b) Data set normalized as monthly hydro generation time

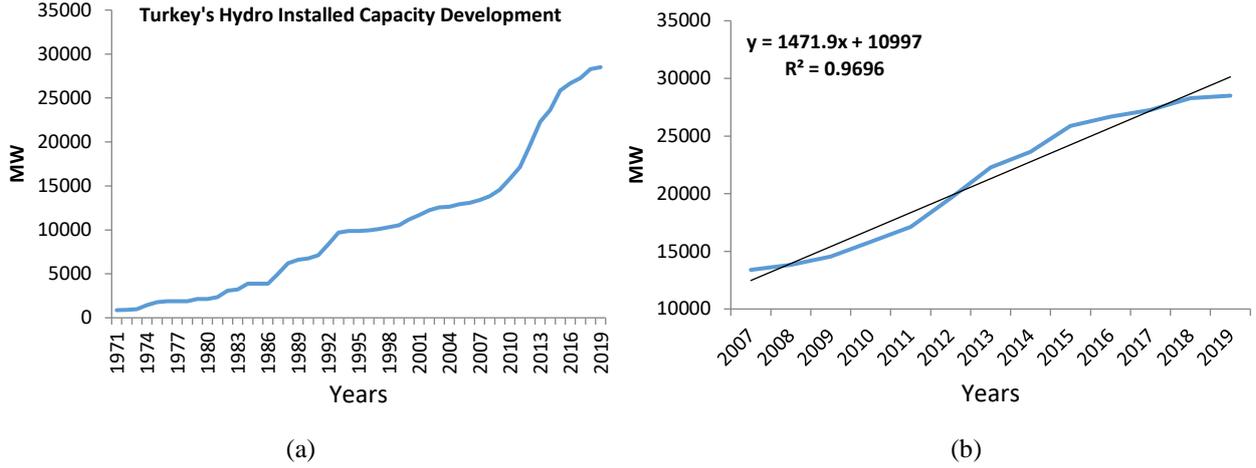

**Figure 9.** a) Hydroelectric installed capacity development, b) Hydroelectric capacity regression model

It is used as two separate data inputs into the LSTM Based hydroelectric system. Long-term, monthly hydroelectric generation covering a 12-year time interval and hydroelectric installed capacity value for these years (Figure 9-a). The fusion of the estimated normalized hydropower generation with the annual hydroelectric generation forecast data monthly is made using the installed capacity obtained from the regression forecast model (Figure 9-b).

**B. LSTM Based Hydro Forecasting Structure**

The block diagram of the LSTM-based deep learning hydroelectric generation system aimed in this study is given in Figure 10. When estimating hydroelectricity generation, first, the generation dataset is standardized by proportioning the annual installed capacity value to the boxed capacity, and after training the LSTM deep network with this generation-based time series, to convert the monthly-based generation forecast values of the next year to the generation values in MWh, the hydroelectric generation board of that year. The capacity value was estimated by the regression model. To make an accurate estimation of the regression model used here high $R^2$ monovalent annual capacity value of the last five years to achieve linear function it is used.

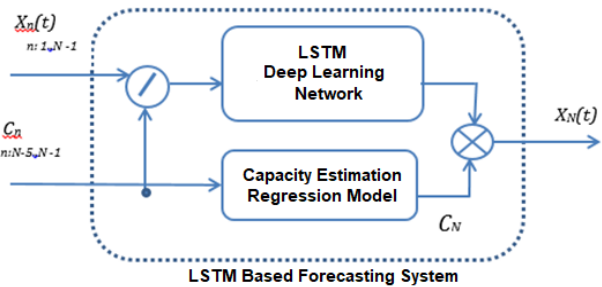

**Figure 10.** Block diagram structure of the forecasting system

Data set used in this study, the monthly value by Turkey's hydroelectric transmission system operator in the country (Transmisio System Operator-TSO) has established is provided from open sources. This set of data to work in January 2007-is to comprise the months of December 2018 includes total of 144 months hydropower production information. This data set is divided into three parts and made to work with three LSTM estimated model. Detailed description of the data set are shown in Table 1.

**Table 1.** Characteristics of the dataset used

| Data set | Time Range/Feature | Parameter |
|---|---|---|
| Hydropower generation time series (MWh) | 01.2007 – 12.2018 | n*t number of months 144 pieces |
| Data time series 72 months: 120 months: 144 months: | 01.2007 – 12.2012 01.2007 – 12.2016 01.2007 – 12.2018 | $X_n(t)$, production value in t.month of n.year |
| Installed Capacity (MW) | 2007 - 2018 | Annually: 12 pieces |

Data Preparation and Calculation of Final Forecast Production value;

1- Since the total installed capacity providing hydraulic electricity generation is different every year; To standardize annual hydraulic electricity production monthly per MW capacity for all years, monthly hydroelectric production (MWh) is proportioned to the final total installed capacity value (MW) of that year, and time curves based on the number of hydroelectric production hours produced for each month are obtained.
2- Since the annual hydroelectricity was standardized using the annual installed capacity value, the estimated year's values are multiplied by the estimated year's total installed capacity value to obtain the monthly hydroelectricity production (MWh) estimated values.

$$EMP = EPT * IP \qquad (8)$$

Here,
EMP: Estimated monthly production (unit: MWh)
EPT: Estimated Production time (unit: h)

3- The graphs of the estimation results made with the production data set of the estimation system with the production time series of 6, 10, and 12 years are given. Detailed analysis of the results and performance evaluation is given in the next section.

As the first LSTM-1 Model; Covering the year 2007 - 2012 72 monthly time series of hydroelectric production was discussed. The results obtained from the model and the results obtained using 100, 200, and 400 layer long short-term memory structures are presented in Figure 12.

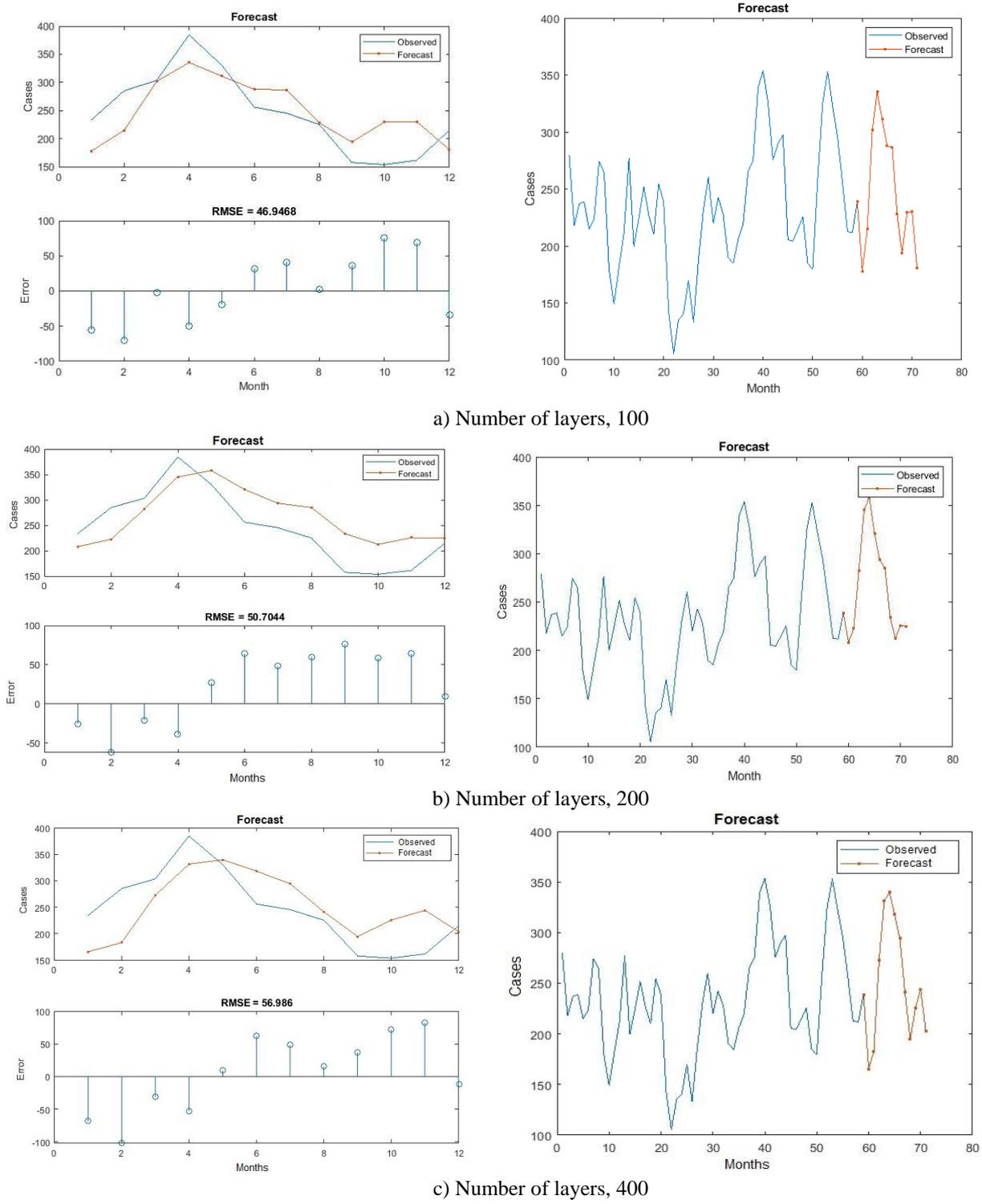

a) Number of layers, 100

b) Number of layers, 200

c) Number of layers, 400

**Figure 12.** Forecast results, error graph and RMSE values for different models using 72-month data series

As the second LSTM-1 Model; The 120-month hydroelectric generation time series covering the years 2007 - 2016 is discussed. The results obtained from the model and the results obtained using 100, 200, and 400 layer long short-term memory structures and their RMSE values are given in Figure 14.

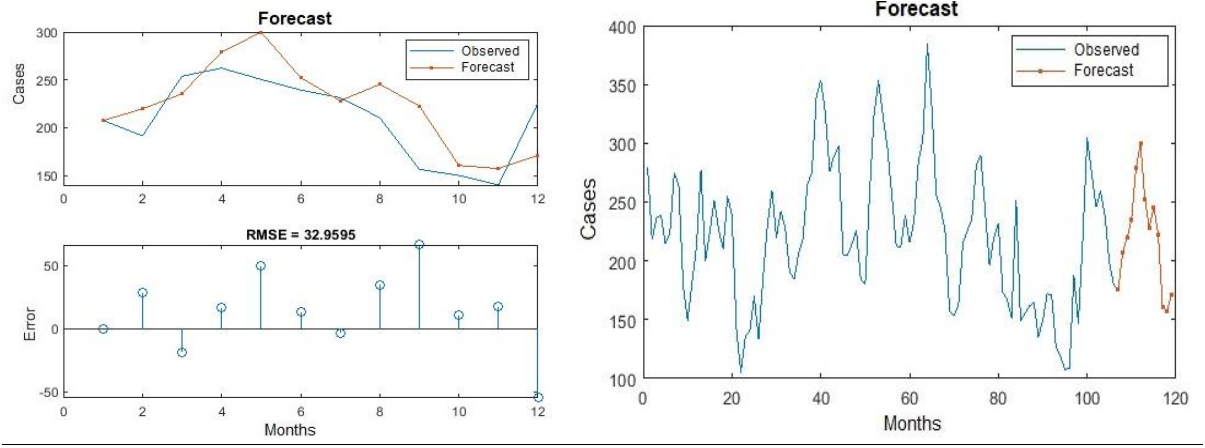

a) LSTM network layer number: 100

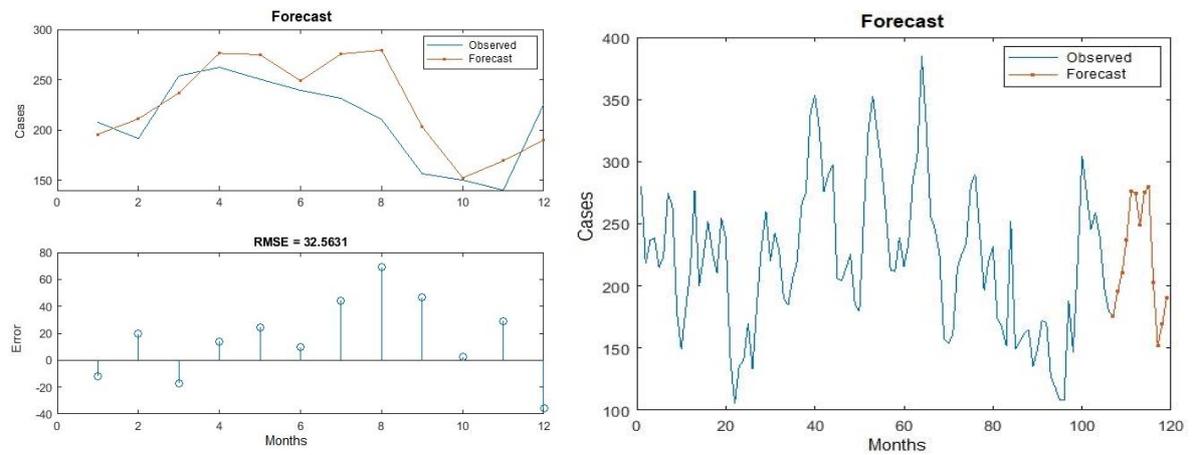

b) LSTM network layer number: 100

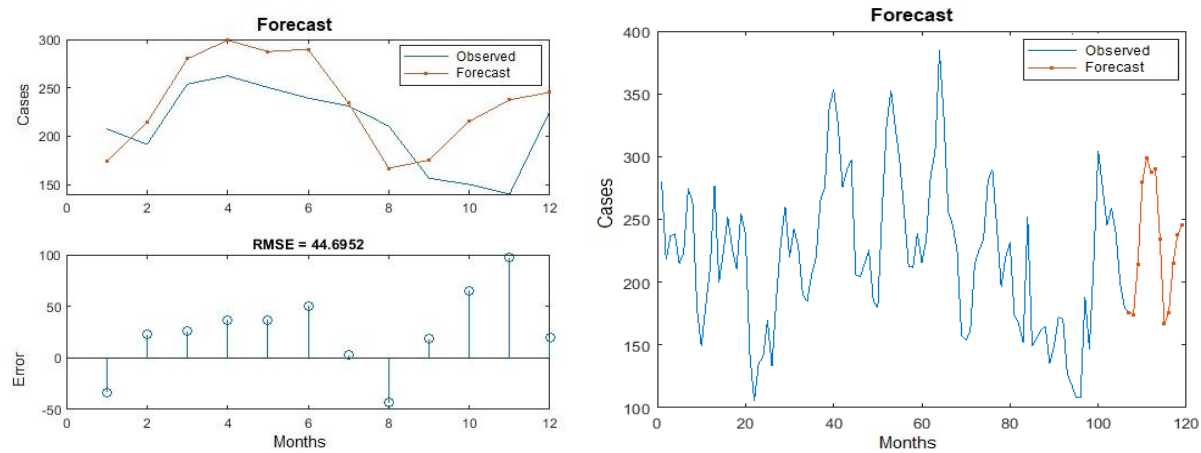

c) LSTM network layer number: 400

**Figure 14.** Forecasting results, error graph and RMSE values for different models a) 100 layers, b) 200 layers, c) 400 layers using a 120-months production data series

As the third LSTM-1 Model; The 144-month hydroelectric generation time series covering the years 2007 - 2018 is discussed. The graph showing the monthly forecast and real values obtained from the model using long short-term memory structures and the result graphics for the 100, 200 and 400 layer structures attached to the time series of the monthly forecast production are given in Figure 16.

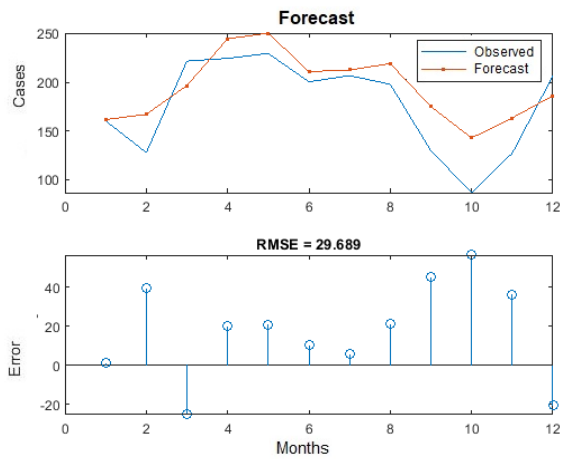
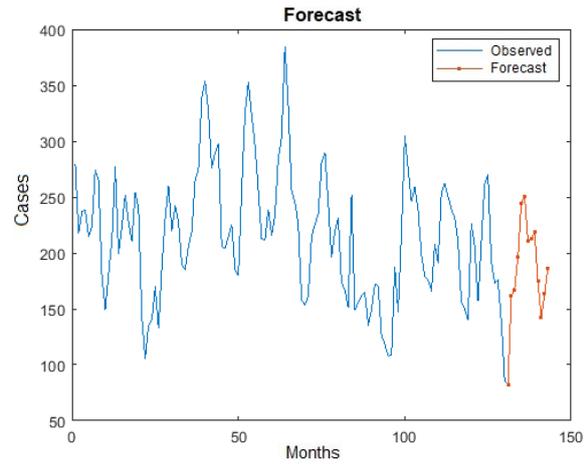

a) LSTM network layer number: 100

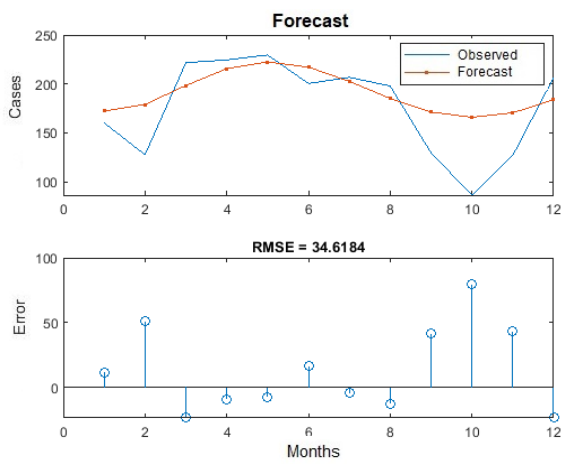
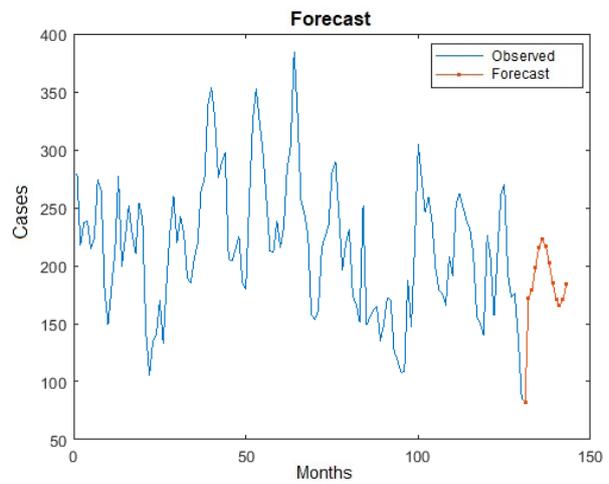

a) LSTM network layer number: 200

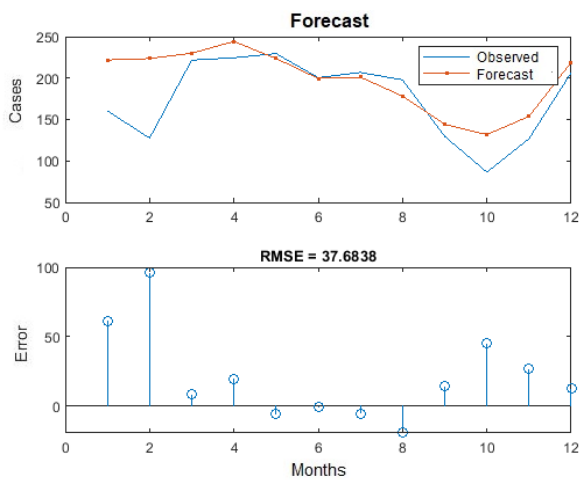
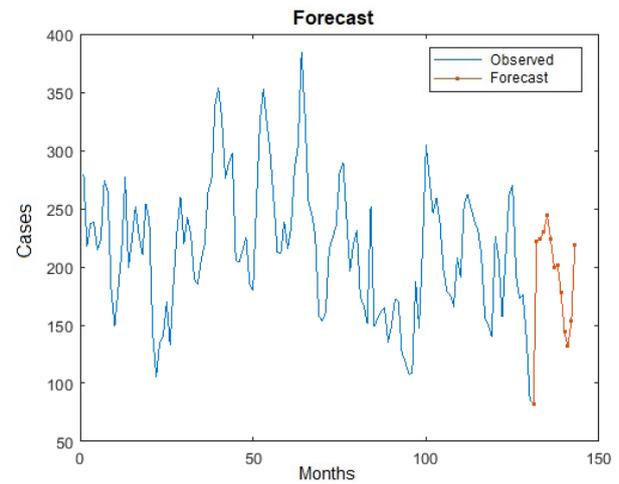

c) LSTM network layer number: 400

**Figure 16.** Forecasting results, error graph and RMSE values for different models a) 100 layers, b) 200 layers, c) 400 layers using a 144-months production data series

RMSE and MAPE values were used to compare the prediction accuracy performances of different long short-term memory structures of models belonging to different time series and to measure the results obtained.

## 5. RESULTS AND DISCUSSION

Well-known evaluation criteria in the literature were used to measure the performance of LSTM-based deep learning networked prediction models. These are Mean Absolute Percent Error (MAPE) and Root Mean Square Error (RMSE). The RMSE is a quadratic scoring rule that also measures the mean magnitude of error and is defined as follows.

$$RMSE = \sqrt{\frac{1}{N}\sum_{i=1}^{N}(v_{estimated} - v_{real})^2} \quad (9)$$

Average absolute percent error (MAPE) is a widely used statistical method that measures how close the forecast result made by a forecasting system is to the truth. It measures accuracy as a percentage. The MAPE method has a high sensitivity to the scale size of the system and is not recommended for use with low volumes of data. With a low-volume data set, the desired level of MAPE values shows the robustness of the prediction system. Also, MAPE will take extreme values when the actual value is not zero, but rather small. It is defined as follows

$$MAPE = \left(\frac{1}{N}\sum\frac{|v_{real} - v_{estimated}|}{|v_{real}|}\right) * 100 \quad (10)$$

In this study, MAPE and RMSE values were used to select the training algorithm of the model with the smallest estimation error by using the hydroelectric monthly generation data time series.

K-fold validation is one of the common methods used to segment the dataset for the training and evaluation of prediction models. To measure the accuracy of the developed LSTM-based deep learning-based prediction system, a 3-fold verification method was used on the dataset. In this process, the data were first divided into 3 different sections and each time the pieces were tested with the annual production values, which included the 12-month value series that followed, while the remaining sections were used for training.

**Table 2.** RMSE and MAPE values obtained for the designed LSTM Models

| Model No | Layer Number | Time Interval | Trainnig Part | Testing Part | RMSE Annual | RMSE Monthly Avg | MAPE Annual | MAPE Monthly Avg (%) |
|---|---|---|---|---|---|---|---|---|
| 1 | 100 | 72 ay | 60 ay | 12 ay | 46,946 | 3912.16 | 0.1907 | 1.58 |
| 1 | 200 | 72 ay | 60 ay | 12 ay | 50,704 | 4225.33 | 0.2173 | 1.81 |
| 1 | 400 | 72 ay | 60 ay | 12 ay | 56,986 | 4748.83 | 0.2251 | 1.88 |
| 2 | 100 | 120 Ay | 108 Ay | 12 ay | 32,959 | 2746.58 | 0.1311 | 1.09 |
| 2 | 200 | 120 Ay | 108 Ay | 12 ay | 32,563 | 2713.58 | 0.1347 | 1.12 |
| 2 | 400 | 120 Ay | 108 Ay | 12 ay | 44,695 | 3724.58 | 0.2031 | 1.69 |
| 3 | 100 | 144 Ay | 132 Ay | 12 ay | 29,689 | 2474.08 | 0.1819 | 1.52 |
| 3 | 200 | 144 Ay | 132 Ay | 12 ay | 34,618 | 2884.83 | 0.2095 | 1.75 |
| 3 | 400 | 144 Ay | 132 Ay | 12 ay | 37,683 | 3140.25 | 0.1938 | 1.62 |

In the study, LSTM networks with 100, 200 and 400 layers were used in 3 different models, each of which was operated with monthly production data of 72 months, 120 months and 144 months. According to Table 2, which contains the RMS and MAP values of the results obtained in the study was 120 months (10 years) hydroelectric time data of application of the 100 layer LSTM model of the annual total 0.1311 (13.1%) and monthly average dispersion 1:09% in terms of prediction accuracy by the MAP value appears to be the highest model. Therefore, within the scope of this study, time data covering at least 120 months of production is recommended to create a hydroelectric forecast model. Additionally, as in this model year with a total of 29.689 monthly distributions at 2474.08 and the RMS value of 144 months (12 years) has been the best result for hydroelectric production data that is used when 100 LSTM layered model data.

When the results obtained from this study are evaluated, in the proposed Long Short-Term Memory Networks (LSTM) Based Hydropower Generation Forecasting System, when the monthly hydroelectric generation data length is selected as at least 120 months and the Long Short-Term Memory Network is 100 layers, it is possible to forecast hydropower generation for the next 12 months (1 year). The results is considered that the prediction accuracy performance decreases in other models where the time data length is more minimal or in the models where the number of layers is high compared to the time data length.

## 6. CONCLUSION AND RECOMMENDATION

The evaluation of renewable energy sources in electricity generation mainly depends on local environmental and meteorological conditions such as temperature and precipitation-runoff rates. Therefore, the expected power generation fluctuates greatly, making the calculation and forecasting of the supply to the power grid difficult. Hydroelectric power is currently the most important renewable resource contributing to the electricity supply in Turkey, and its contribution is expected to increase further in the future. Accurate estimation of energy production is a crucial issue for the current power management process. Therefore, the model in the present study represents a major improvement in this field and a contribution to the existing literature. This study focuses on the potential of using deep learning LSTM to forecast annual hydroelectric power demand on monthly basis. For this, an estimation system based on annual hydroelectricity installed capacity development with monthly hydroelectricity production time data has been proposed. The results of the research show that LSTM provides a strong architecture for forecasting hydropower generation in both medium and long-term forecasts.

Statistical measures root mean square error (RMSE) and mean absolute percentage error (MAPE) was used to analyze the performance of three learning algorithms based on 72-month, 120-month and 144-month generation hydroelectric generation data used in the study. The best estimation was obtained in the 100-layer LSTM model using 120 months (10 years) hydroelectric generation time data with the lowest MAPE percentage, and the 100-layer LSTM models using 144 months (12 years) hydroelectric generation data time data with the lowest RMSE value . The more data available, the more accurate the predictions are, the more accurate the results are demonstrated here. However, if a similar study is conducted on a larger training set covering longer years, it is predicted that the algorithms used for the proposed estimation system will yield better results in their performance.

**Conflict of Interest**
The authors declare that there is no conflict of interest.